\documentclass[conference]{IEEEtran}
\IEEEoverridecommandlockouts
\pdfoutput=1
\usepackage{cite}
\usepackage{amsmath,amssymb,amsfonts}
\usepackage{mathrsfs}
\usepackage{algorithmic}
\usepackage{graphicx}
\usepackage{textcomp}
\usepackage{xcolor}
\usepackage{array}
\usepackage{multirow,multicol}
\usepackage[caption=false,font=footnotesize]{subfig}
\usepackage{hyperref}
\usepackage[final]{pdfpages}
\usepackage{arydshln}
\def\BibTeX{{\rm B\kern-.05em{\sc i\kern-.025em b}\kern-.08em
    T\kern-.1667em\lower.7ex\hbox{E}\kern-.125emX}}
\begin{document}
\newcommand{\norm}[1]{\left\lVert#1\right\rVert}
\title{Towards a Deep Unified Framework for Nuclear Reactor Perturbation Analysis
{}
}
\author{\IEEEauthorblockN{ Fabio De Sousa Ribeiro* \\ and Francesco Caliv\'a*}
\IEEEauthorblockA{University of Lincoln, UK\\
\{fdesousaribeiro,fcaliva\}\\@lincoln.ac.uk\\
*Both authors contributed equally}
\and
\IEEEauthorblockN{Dionysios Chionis \\ and Abdelhamid Dokhane}
\IEEEauthorblockA{Paul Scherrer Institute\\
Villigen, Switzerland \\
\{dionysios.chionis,\\abdelhamid.dokhane\}@psi.ch}
\and
\IEEEauthorblockN{Antonios Mylonakis\\ and Christophe Demazi\`ere}
\IEEEauthorblockA{Chalmers University of\\
Technology, Sweden \\
\{antmyl,demaz\}\\@chalmers.se}
\and
\IEEEauthorblockN{Georgios Leontidis\\and Stefanos Kollias}
\IEEEauthorblockA{University of Lincoln\\
Lincoln, UK \\
\{gleontidis,skollias\}\\@lincoln.ac.uk}
}
\maketitle

\begin{abstract}
In this paper, we take the first steps towards a novel unified framework for the analysis of perturbations in both the Time and Frequency domains. The identification of type and source of such perturbations is fundamental for monitoring reactor cores and guarantee safety while running at nominal conditions. A 3D Convolutional Neural Network (3D-CNN) was employed to analyse perturbations happening in the frequency domain, such as an absorber of variable strength or propagating perturbation. Recurrent neural networks (RNN), specifically Long Short-Term Memory (LSTM) networks were used to study signal sequences related to perturbations induced in the time domain, including the vibrations of fuel assemblies and the fluctuations of thermal-hydraulic parameters at the inlet of the reactor coolant loops. 512 dimensional representations were extracted from the 3D-CNN and LSTM architectures, and used as input to a fused multi-sigmoid classification layer to recognise the perturbation type. If the perturbation is in the frequency domain, a separate fully-connected layer utilises said representations to regress the coordinates of its source. The results showed that the perturbation type can be recognised with high accuracy in all cases, and frequency domain scenario sources can be localised with high precision.
\end{abstract}

\section{Introduction}
For over half a century, the nuclear industry has primarily focused on the technological evolution of reliable nuclear power plants for the production of electricity. By monitoring nuclear reactors while running at nominal conditions, it is possible to gather valuable insight for early detection of anomalies. Various types of fluctuations can be caused by the turbulent nature of flow in the core, mechanical vibrations within the reactor, coolant boiling and stochastic character of nuclear reactions, among other factors. These fluctuations are often referred to as neutron noise $\delta X(\mathbf{r},t)$, which is measured as in (\ref{eq: noise delta X}), where $X(\mathbf{r},t)$ represents the signal and $X_{0}(\mathbf{r},t)$ its trend. Both are a function of two variables: $\mathbf{r}$ the spatial coordinate within the core, and $t$ time.
\begin{equation}
\label{eq: noise delta X}  
  \delta X(\mathbf{r},t) = X(\mathbf{r},t) - X_{0}(\mathbf{r},t)
\end{equation}
\begin{figure}[t]
\centering
\includegraphics[trim={10 5 25 10},clip,width=0.9\columnwidth]{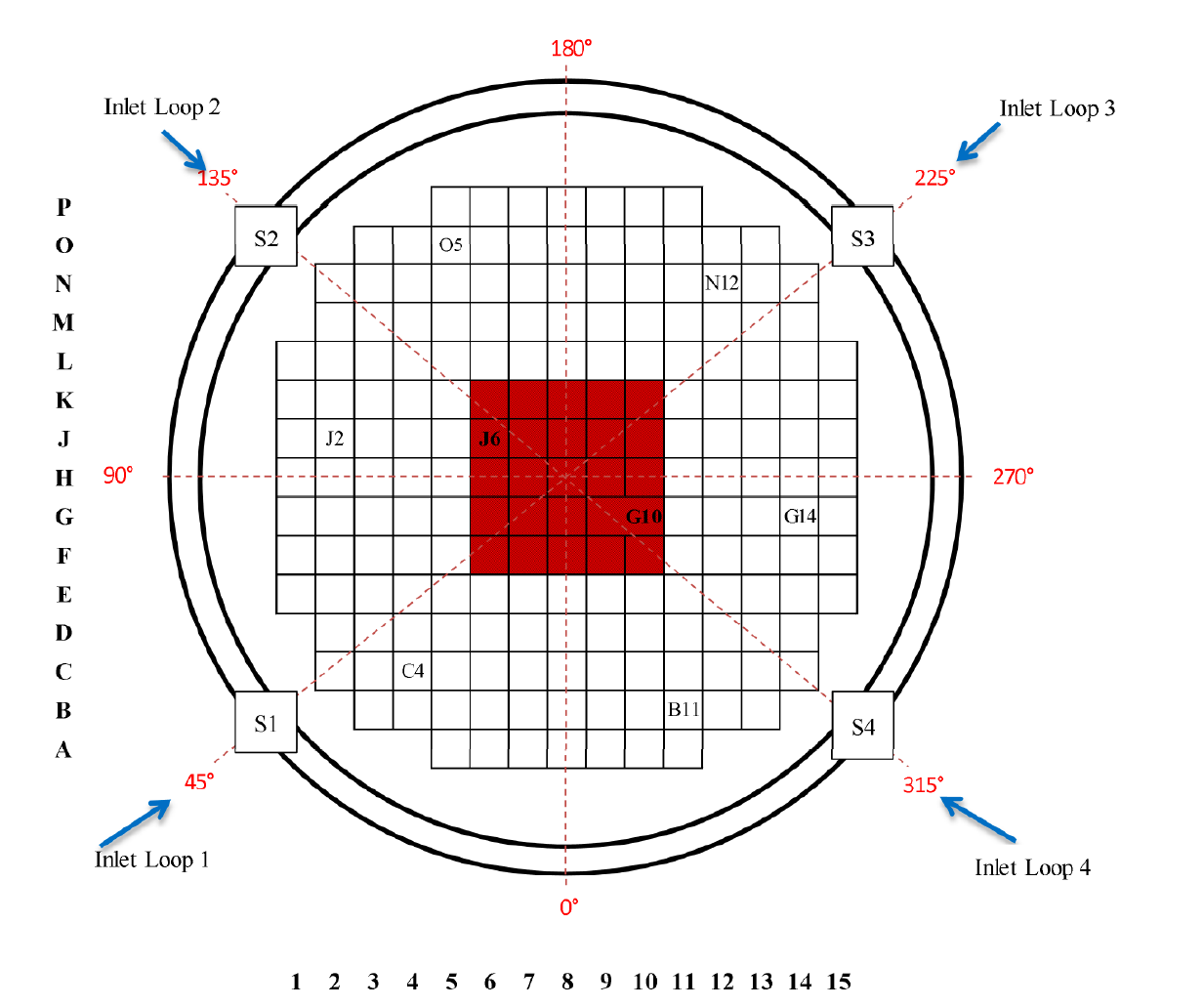}
\caption{Illustrative radial view of the nuclear reactor core model utilised in Simulate-3K. Each letter and number pairing denotes an in- or ex-core signal detector, and each grid square represents a fuel assembly. The red central zone represents a $5\times5$ cluster of fuel assemblies that vibrates synchronously in the $x$ direction. The calculated neutron noise distribution was utilised in our deep learning based analysis of perturbations in the Time Domain.}
\label{fig:reactor_core_section}
\end{figure}
With detailed descriptions of reactor geometry, physical perturbations and probabilities of neutron interactions within the core -- by assuming a particular reactor transfer function (i.e. Green's function) -- one can simulate how fluctuations affect the neutron flux in the time or frequency domain. Different types of perturbations can then be applied in order to estimate and study the induced neutron noise, as to solve the \textit{forward problem}. Intuitively, the \textit{backward problem}, also known as \textit{unfolding}, consists of localising the perturbation origin and can only be carried out if the reactor transfer function is inverted. Solving the unfolding problem is therefore non-trivial as measurements of the induced neutron noise are not available at every position inside the reactor core, due to a limited number of in- and ex-core sensors available. 
\begin{figure*}[!ht]
\centering
\begin{tabular}{ccc}
\subfloat{\includegraphics[width=0.63\columnwidth]{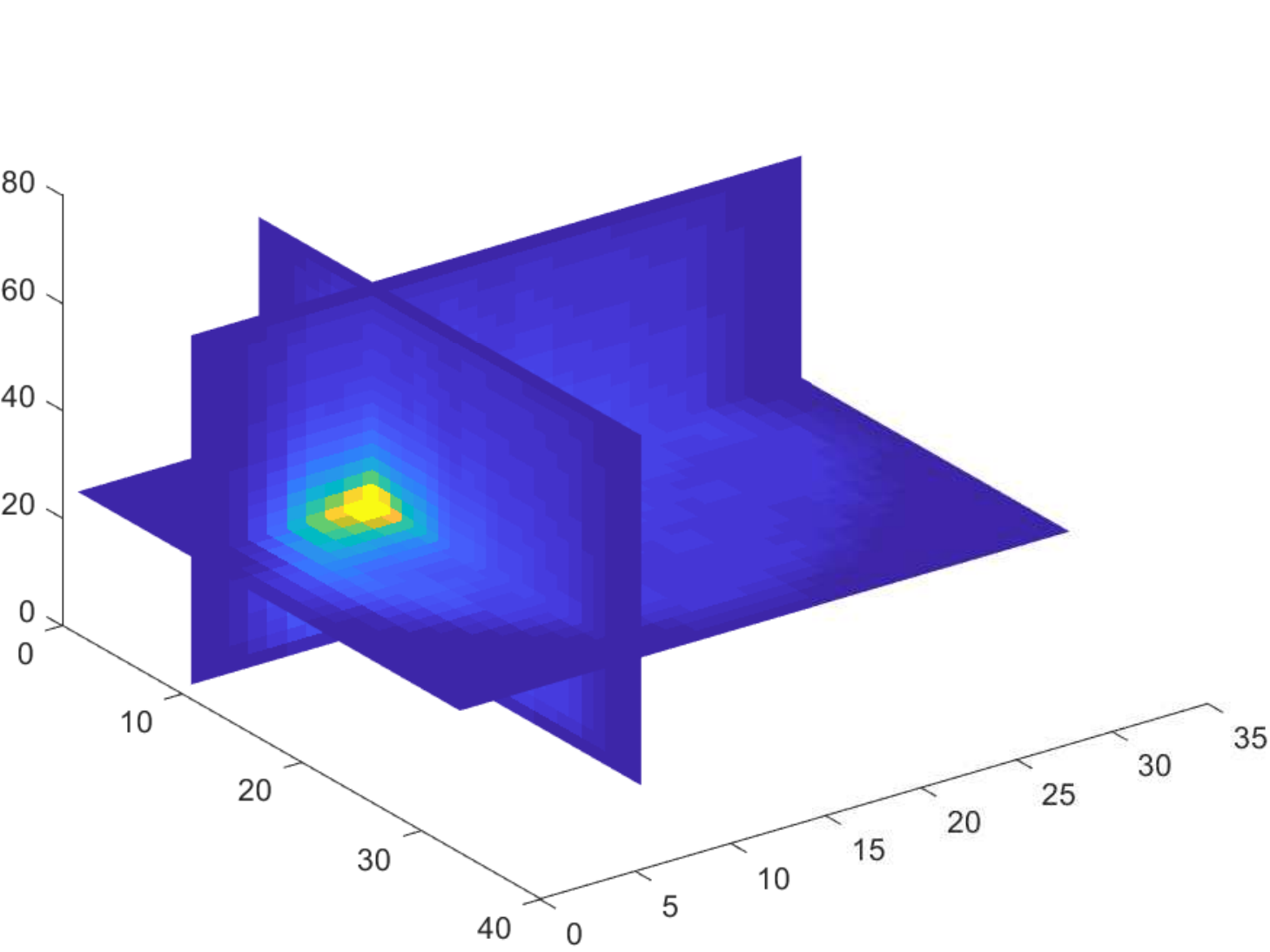}} & \subfloat{\includegraphics[width=0.63\columnwidth]{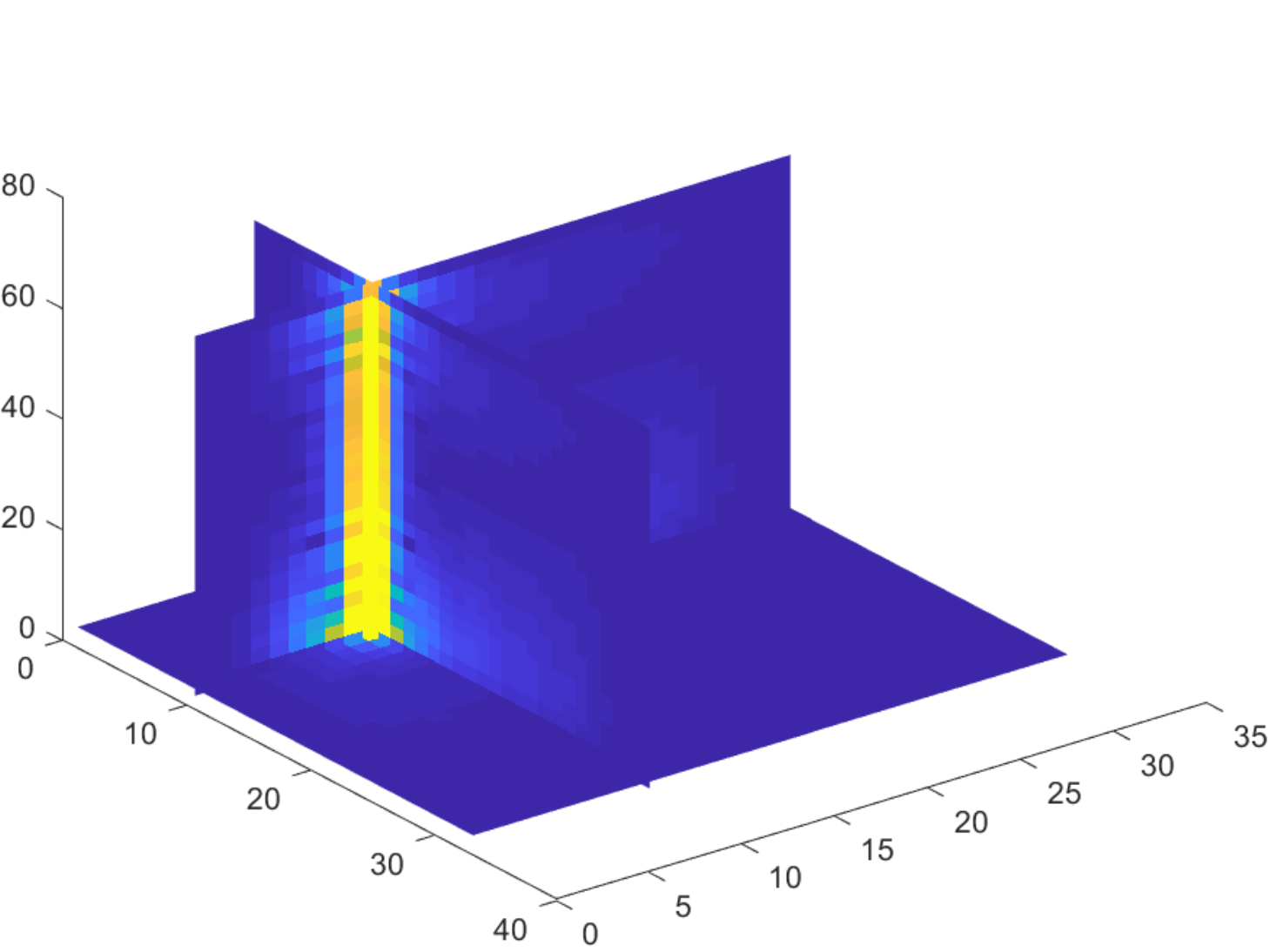}} &
\subfloat{\includegraphics[width=0.63\columnwidth]{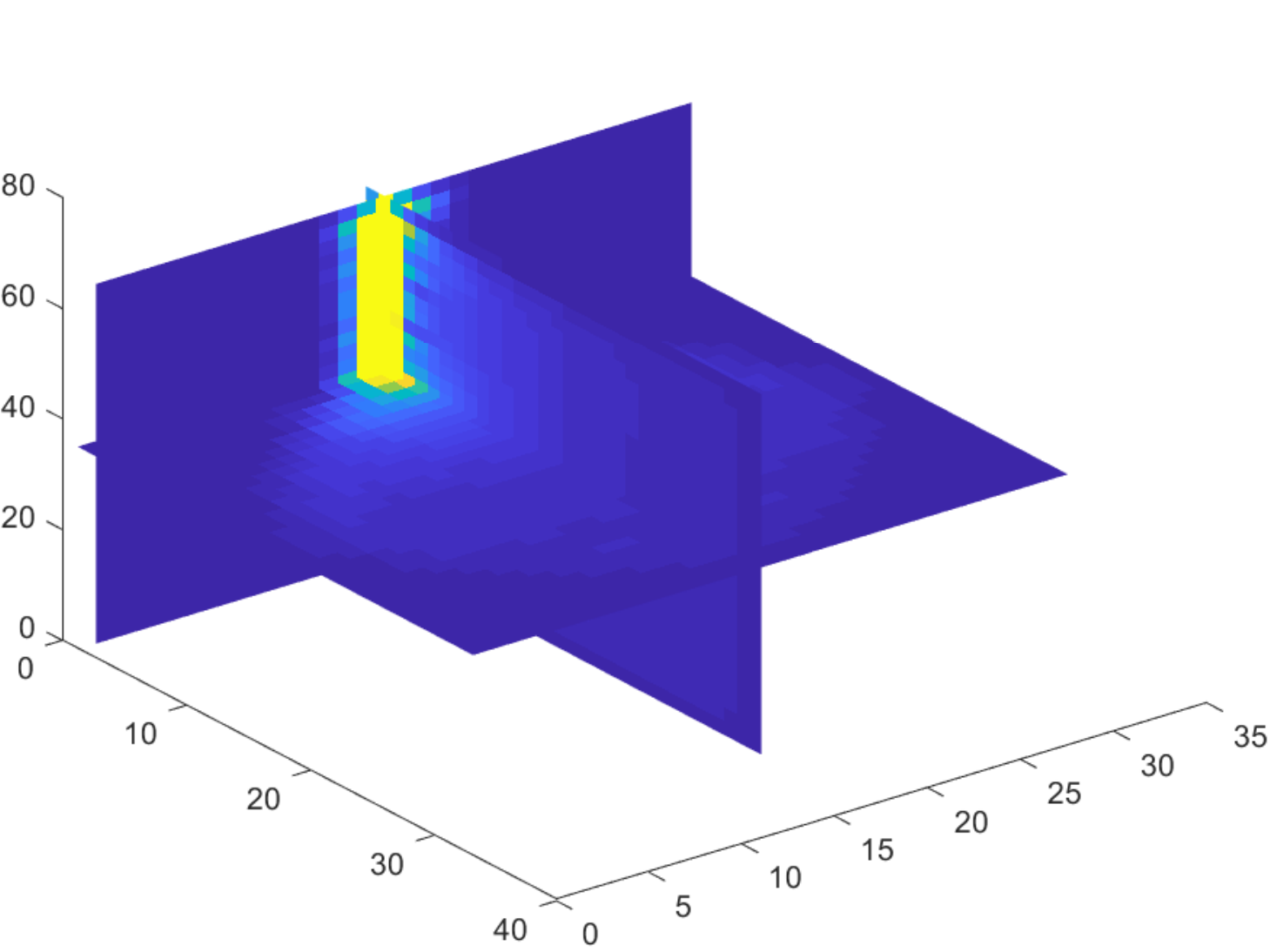}} 
\end{tabular}
\caption{Examples of induced neutron noise types. From \textbf{Left} to \textbf{Right}, the responses to \textit{Localised}, \textit{Propagating} type $1$ and $2$ perturbations are illustrated.}
\label{fig:perturbations}
\end{figure*}
In this work, a novel method to unfold nuclear reactor signals pertaining to the localisation of different types of perturbations is proposed. This is achieved by extending and improving previous research on the application of deep learning techniques to detect anomalies in nuclear reactors \cite{caliva2018deep}.

\section{Related Work}
Fault detection in nuclear reactors has been the focus of a few recent studies. \cite{Jin2011anomaly} proposed a pattern recognition framework to detect anomalies based on symbolic dynamic filtering of time series data. \cite{ZAFERANLOUEI2010813} predicted critical heat flux by ways of Adaptive Neuro-Fuzzy Inference Systems (ANFIS). \cite{LI201792} monitored sensors by utilising auto-associative kernel regression and sequential probability ratio tests. \cite{SANTOSH2009759} collected reactors parameters and implemented an artificial neural network (NN) based system to diagnose transients. \cite{JAMIL2016433} proposed a nuclear reactor fault detector based on the combination of principal component analysis and fisher discriminant analysis. Deep learning has recently shown to be effective in a variety of safety-critical fields spanning from signal analysis, to computer vision applications including medical imaging and text recognition (\cite{krizhevsky2012imagenet,yosinski2014transferable,karpathy2014large,kollias2017adaptation,kollias2017deep,de2018adaptable,de2018end}). In \cite{chen2018deep}, a Convolutional NN (CNN) and Na\"ive-Bayes data fusion scheme was proposed to detect fractures in plant components by way of individual video frame analysis. In \cite{hosseini2017neutron}, a dynamic Galerkin Finite Element Method-based simulator was applied to calculate the frequency domain neutron noise distribution of the VVER-1000 core reactor. Subsequently, an ANFIS was employed to localise the induced neutron noise source. Conversely, in a recent work by~\cite{caliva2018deep}, the induced neutron noise was simulated by using CORE-SIM, at different perturbations strengths and frequencies. A CNN was employed to localise the origin of frequency domain neutron noise perturbations in nuclear reactor signals. This was achieved by spatially splitting the complex signal volumes into $12$ or $48$ individual blocks, each pertaining to a different class. A classification task was then formulated, followed by a combination of $k$-means and $k$-NN based analysis of extracted latent variables, enabling a finer unfolding resolution. Although the results were promising, an unfounded conversion of complex signal volumes for use in conventional CNNs led to unnecessary loss of spatial information. To address this limitation, in this work we propose a new bespoke 3D CNN model for multi-task perturbation unfolding regression and type classification. Additionally, we extend our analysis to time-domain simulated signals regarding vibrating fuel assemblies and/or fluctuations in thermal-hydraulic parameters (e.g. inlet coolant flow/temperature).

\section{The Examined Scenarios}
\label{sec: scenarios}
\subsection{Frequency Domain} 
In this study, CORE-SIM~\cite{demaziere2011core} was employed to model the induced neutron noise, in a Pressurised Water Reactor (PWR), under two scenario settings: \textit{Absorber of Variable Strength} and \textit{Propagating Perturbation} in the frequency domain. During the forward problem, the reactor transfer function, which is considered to be the Green's function of the system, captures the response of the induced fluctuations in neutron flux. The effect of a perturbation can be assessed from any spatial point within the reactor core, provided that there exists a one-to-one relationship between every possible location where a perturbation is located and the position where the neutron noise is measured. The latter is described as
\begin{equation} \label{sec:Introduction - subsec: related work - basic scenarion - eq: induced neutron noise}
\delta \phi(\mathbf{r},\omega) = \int_{V} G(\mathbf{r},\mathbf{r_{p}},\omega) \delta S(\mathbf{r_{p}}) d\mathbf{r_{p}},
\end{equation}
where the core transfer function is integrated across the whole core reactor volume $V$, whereas $\mathbf{r_{p}}$ and $\omega$ refer to the source and the angular frequency of the perturbation respectively. For more details, please refer to the official CORE-SIM user manual~\cite{demaziere2011core,demaziere2011user}. Diffusion theory was applied to perform a low-order approximation of the angular moment of the neutron flux. The energy of the system was discretised with a two-energy group formulation: one with a high and one low energy spectrum, henceforth referred to as the \textit{Fast} and the \textit{Thermal} groups respectively.
\subsubsection*{Absorber of Variable Strength} 
In this scenario (\textit{Localised}, see Fig.~\ref{fig:perturbations}), the thermal macroscopic absorption cross-section was perturbed at three different frequencies $0.1$, $1$ and $10~\text{Hz}$, altering the absorption of thermal neutrons. This perturbation type can be considered as localised at a specific source location. A PWR with a radial core of size $15\times15$ fuel assemblies (FA) was modelled, using a volumetric mesh with $32\times32\times26$ voxels.
\subsubsection*{Propagating Perturbation} 
In these scenarios (\textit{Propagating} type $1$ and $2$, see Fig.~\ref{fig:perturbations}), fuel assemblies were also perturbed at $0.1$, $1$ and $10~\text{Hz}$, at which the fluctuations in neutron noise were modelled. Propagating perturbations were located either at the core inlet and transported upwards with the coolant starting from the lowest level of the core (type $1$); or within the core and propagated along the fuel assembly's cross-section, by means of the coolant flow (type $2$). See Fig.~\ref{fig: fuel_assembly} for intuition. Identical mesh specifications to the \textit{Absorber of Variable Strength} scenario were adopted.
\subsubsection*{Combined Perturbations}
\label{subsec: combined}
In this scenario, combinations of the aforementioned perturbation types can occur simultaneously at different locations in the reactor. However, no more than one instance per perturbation type can occur at any given time. 
\subsubsection*{Data Pre-processing}
\label{subsec: data-preprocessing}
The complex signals are a 3D representation of the distribution of the induced neutron noise, including \textit{Fast} and \textit{Thermal} neutron groups. They are distributed in the form of voxels of size $32\times32\times26$, each containing a perturbation located at a specific coordinate location $i,j,k$ (considered as the label of our regression task). The dataset is comprised of $19552$ (\textit{Absorber of Variable Strength}) and $752$ (\textit{Propagating} type $1$ and $2$) instances per frequency ($0.1$, $1$ and $10~\text{Hz}$). Furthermore, the signal was corrupted by obscuring parts (set values to zero) at random in order to emulate fewer available sensor measurements. Two versions of obscured data were generated with channel-wise repeated masks of size $32\times32\times26$. Each $32\times32$ mask was generated by randomly selecting $5\%$ and $20\%$ of measurements respectively, and setting remaining values to zero. As previously alluded to, a given reactor signal is composed of $2$ types of responses, \textit{Fast} and \textit{Thermal}, each comprised of amplitude and phase. Resulting in a total of $4$ components of size $32\times32\times26$, which we concatenated into a $64\times64\times26$ volume, zero-padded to $64\times64\times32$ for convenience. 
\begin{table}[]
\caption{Synchronised vibration of a $5\times5$ fuel assemblies central cluster.}
    \label{table: time_scenarios1-4}
    \centering
    \setlength{\extrarowheight}{2pt}
    \begin{tabular}{c|cccc}
    \hline  \hline 
    Scenario & Perturbation & Frequency & Amplitude & ID \\
    \hline
    \multirow{2}{*}{1}     & $5\times5$ cluster FAs & WN & $1~mm$ & 1 0 0 0  \\
         & $5\times5$ cluster FAs & WN & $0.5~mm$ & 1 0 0 0  \\
         \hline
    \multirow{2}{*}{2}     & $5\times5$ cluster FAs   & $1~\text{Hz}$ & $1~mm$ & 0 1 0 0  \\
     & $5\times5$ cluster FAs   & $1~\text{Hz}$ & $0.5~mm$ & 0 1 0 0  \\
    \hline \hline
    \end{tabular}
\end{table}
\begin{table}[]
\caption{Synchronised perturbation of coolant thermal-hydraulic parameters.}
    \label{table: time_scenarios5-6}
    \centering
    \setlength{\extrarowheight}{2pt}
    \begin{tabular}{c|cccc}
    \hline  \hline 
    Scenario & Perturbation & Frequency & Amplitude & ID \\
    \hline
    3    & temperature  & random & $\pm 1^{\circ}C$ & 0 0 1 0  \\
    4    & flow  & random & $\pm 1\%$ & 0 0 0 1  \\
    \hline \hline
    \end{tabular}
\end{table}
\subsection{Time Domain}
Simulate-3K (S3K) was utilised to model fuel assemblies cluster vibrations, including fluctuations in thermal-hydraulic parameters between the coolant loops, on a model of the four-loop Westinghouse PWR mixed core, utilised in \cite{kozlowski2003oecd}. The system operating conditions were close to those used in the frequency domain experiments. For more details with regard to S3K, the reader is invited to refer to the manual~\cite{grandi2006simulate}. Fig.~\ref{fig:reactor_core_section} depicts a cross-sectional view of the utilised core. The cluster of fuel assemblies is highlighted in red, whereas the coordinates (e.g. B11) identify the location of neutron detectors. Detector-wise, the reactor is comprised of six axial levels and a total of fifty-six detectors: eight located ex-core, identically distributed at two axial levels (level $1$ ($L1$) and level $6$ ($L6$)); forty-eight in-core, equally distributed across the six levels. Every scenario had a duration of $100~s$, sampled with time steps of $0.01~s$, and is briefly explained below.
\begin{table}[]
\caption{Combination of synchronised vibration of a $5\times5$ fuel assemblies central cluster and synchronised perturbation of coolant thermal-hydraulic parameters.}
    \label{table: time_scenarios7-15}
    \centering
    \setlength{\extrarowheight}{2pt}
    \begin{tabular}{c|cc}
    \hline  \hline 
    Scenario & Combined Perturbations &  ID \\
    \hline
    5    & Temperature (5) \& flow (6)   &  0 0 1 1  \\
    6     & $5\times5$ FA (2) \& temperature (5)  & 1 0 1 0  \\
    7     & $5\times5$ FA (1) \& temperature (5)  & 1 0 1 0  \\
    8    & $5\times5$ FA (4) \& temperature (5)  & 0 1 1 0  \\
    9    & $5\times5$ FA (3) \& temperature (5)  & 0 1 1 0  \\
    10    & $5\times5$ FA (2) \& flow (6)  & 1 0 0 1  \\
    11    & $5\times5$ FA (1) \& flow (6)  & 1 0 0 1  \\
    12    & $5\times5$ FA (4) \& flow (6)  & 0 1 0 1  \\
    13    & $5\times5$ FA (3) \& flow (6)  & 0 1 0 1  \\    
    \hline \hline
    \end{tabular}
\end{table}
\begin{figure}[]
\centering
\includegraphics[trim={0 2 0 2},clip,width=0.85\columnwidth]{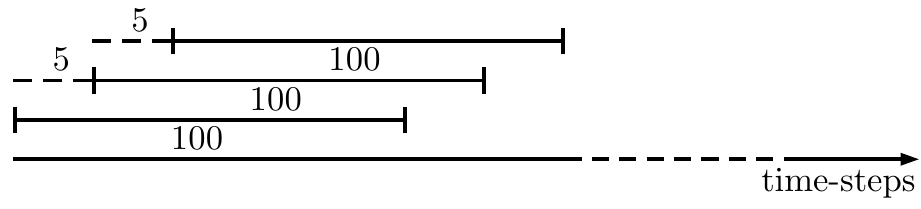}  
\caption{Signal sampling. Signal windows of $100$ time-steps were sampled using sliding windows of stride $5$ time-steps}
\label{fig:signal_sampling}
\end{figure}
\subsubsection{Vibrating central cluster of fuel assemblies}
This perturbation refers to four perturbation instances (see Table~\ref{table: time_scenarios1-4}), in which a cluster of $5\times5$ fuel assemblies is vibrating synchronously in the $x$ direction, following either a white noise signal or a sine wave function $f=0.1~\text{Hz}$, with varying amplitudes in the range of $0.5~mm$, and $1~mm$. ``ID" is a label later utilised to classify different perturbation types. It is worth noting that the first and second rows represent the same scenario, since the applied perturbations are the same but with different amplitude; identical consideration applies to the third and fourth rows. Therefore, two individual scenarios were identified out of the four possible perturbations.
\subsubsection{Perturbation of thermal-hydraulic parameters}
This perturbation refers to two scenarios, in which synchronised fluctuations of inlet coolant temperature between the four coolant loops were induced. As reported in Table~\ref{table: time_scenarios5-6}, the inlet coolant temperature was forced to fluctuate with amplitude of $\pm 1^{\circ}C$ over the mean value of $283.8^{\circ}C$ (third scenario). In the fourth scenario, inlet coolant flow random fluctuations with amplitude of $1$\% over the relative flow ($100$\%) were simulated.
\subsubsection{Combined Perturbations}
Scenarios five to thirteen refer to combinations of previous perturbations associated to the vibration of a $5\times5$ fuel assembly and fluctuations of inlet coolant thermal-hydraulic parameters between the four coolant loops. A detailed description of these scenarios is provided in Table~\ref{table: time_scenarios7-15}. In the column ``Combined Perturbations", the number between brackets links to the Scenario ID reported in Table~\ref{table: time_scenarios1-4} and~\ref{table: time_scenarios5-6}.
\subsubsection*{Data Pre-processing}
\begin{figure}
\centering
\includegraphics[trim={120 40 125 20},clip,width=0.9\columnwidth]{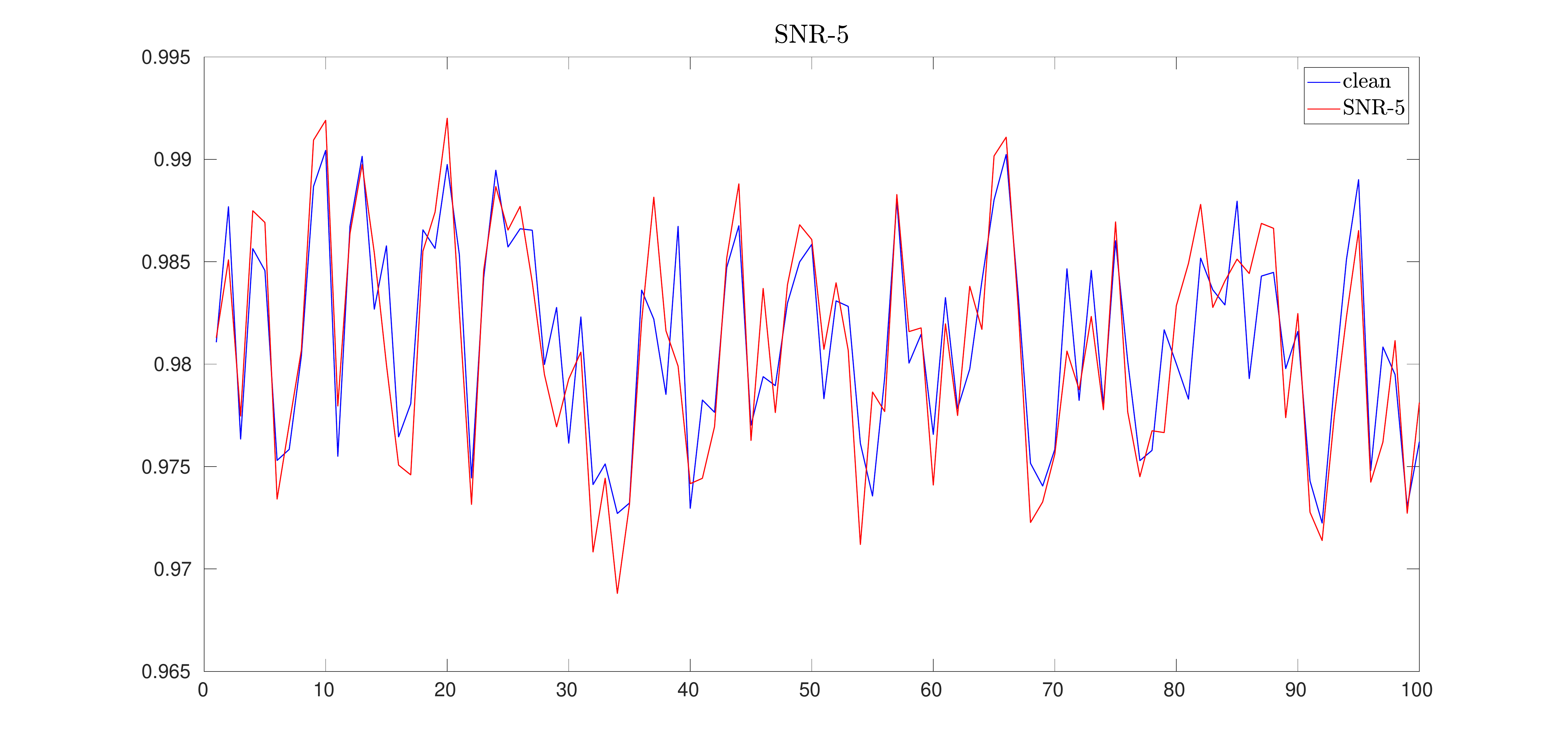} 
\caption{Example of a signal obtained by means of S3K with noise added at SNR$=5$.}
\label{fig:SNR-time}
\end{figure}

Signals produced by S3K are a representation of the neutron flux measured by the in- and ex-core detectors. Taking into account the duration of each applied perturbation and sampling rate, data from each sensor were available in the form of a vector of $10001$ elements. Given the limited amount of data available, it was appropriate to perform data augmentation. To this end, each signal was re-sampled by means of sliding windows as shown in Fig.~\ref{fig:signal_sampling}. Specifically, with sensor measurements over $100~s$ at a sampling rate of $0.01~s$, we get $\mathbf{x} \in \mathbb{R}^{10001}$ signal vectors. These vectors are augmented by means of $100$ time step sliding windows with a stride of $5$ to produce $\mathbf{x} \in \mathbb{R}^{1980\times100}$. Furthermore, the signal was corrupted by the addition of White Gaussian Noise at signal-to-noise ratios (SNR) $10$ and $5$ to study the effect of noisy signals on the performance of our model (Fig.~\ref{fig:SNR-time}).
\section{The Proposed Approach}
\subsection{Frequency Domain} 
Given complex reactor signals in the form of volumetric meshes, it is advantageous to capture spatial information not only in 2D coordinate space ($i$, $j$) but also channel-wise through $k$. This means that knowledge learnt in a particular area of the volume can generalise well to others. The generalisation property of CNNs is crucial, as it allows for a great reduction in the number of parameters when compared to fully-connected (FC) networks, without sacrificing performance. 
However, it is important to state that the signal volumes are not a measure of induced neutron noise over time, but rather a measured response in every ($i$, $j$, $k$) location within the core reactor, in an instant soon after a perturbation is induced. Therefore, the input signal volumes are more closely related to MRI or CT scans rather than videos in terms of data format. Relatedly, 3D CNNs have been used extensively in the medical field for tumour and lesion segmentation, as well as in action recognition tasks to a very good level of success \cite{kamnitsas2017efficient,cciccek20163d,ji20133d,maturana2015voxnet,milletari2016v}. In pursuance of optimal feature extraction in all dimensions of the reactor signal, a bespoke 3D CNN is proposed.
\subsubsection{Convolutional Neural Networks}
Convolutional Neural Networks (CNNs) \cite{lecun1989generalization} perform automatic feature extraction through a series of volume-wise convolutions and feature routing. For each convolutional layer, a resulting set of filters are learnt to capture spatial patterns in given inputs. Deeper CNNs are capable of capturing complex hierarchical concepts, whereby more general and abstract concepts initiate from the stem of the network and become increasingly task specific in the final layers. The convolution operation in CNNs is significantly more efficient than dense matrix multiplication through sparse interactions and parameter sharing. Formally, in 3D CNNs one would compute a pre-activated value of a given unit $n_{i,j,k}^{[\ell]}$ at $(i,j,k)$ position in a 3D feature map of layer $\ell$, by summing the weighted kernel contributions from the previous layer units in $\mathbf{A}^{[\ell-1]}$ as
\begin{equation} \label{3D_convolution}
n_{i,j,k}^{[\ell]}=\sum_{x=0}^{X-1}\sum_{y=0}^{Y-1}\sum_{z=0}^{Z-1}\mathbf{W}_{x,y,z}^{[\ell]}\mathbf{A}_{i+x,j+y,k+z}^{[\ell-1]},
\end{equation}
where $\mathbf{W}_{x,y,z}^{[\ell]}$ is a single learnt weight pertaining to a kernel $\mathbf{W}^{[\ell]}$ of dimensions $X \times Y \times Z$ in layer $\ell$, which is convolved with cells from the previous layer ($\mathbf{W}^{[\ell]}*\mathbf{A}^{[\ell-1]}$). Each feature map $f$ in a given layer $\ell$ has a learnt bias term $b^{[\ell,f]}$, which is added pre non-linearity as 
\begin{equation}
a_{i,j,k}^{[\ell,f]}=\phi(n_{i,j,k}^{[\ell,f]}+b^{[\ell,f]}),
\end{equation}
where $\phi(\cdot)$ is a non-linear activation function such as ReLU$:\rightarrow f(\cdot) = \max(0, \cdot)$ or the logistic sigmoid.
\begin{table}[]
\caption{3D-CNN architecture for Frequency Domain perturbation type classification and source regression.}
    \label{table: 3D-CNN_arch}
    \centering
    \setlength{\extrarowheight}{2pt}
    \begin{tabular}{c|c|c}
    \hline  \hline
    \multicolumn{3}{c}{Input Size:      64$\times$64$\times$32$\times$1}  \\
    \hline
    Conv-BN-ReLU   & 3$\times$3$\times$3@64  & 64$\times$64$\times$32$\times$64  \\
    \hline
    MaxPool    & 2$\times$2$\times$2  & 32$\times$32$\times$16$\times$64  \\
    \hline
    Conv-BN-ReLU    & 1$\times$1$\times$1@32  & 32$\times$32$\times$16$\times$32  \\
    \hline
    Conv-BN-ReLU    & 3$\times$3$\times$3@128  & 32$\times$32$\times$16$\times$128  \\
    \hline
    MaxPool    & 2$\times$2$\times$2  & 16$\times$16$\times$8$\times$128  \\
    \hline
    Conv-BN-ReLU    & 1$\times$1$\times$1@64  & 16$\times$16$\times$8$\times$64  \\
    \hline
    Conv-BN-ReLU    & 3$\times$3$\times$3@256  & 16$\times$16$\times$8$\times$256  \\
    \hline
    MaxPool    & 2$\times$2$\times$2  & 8$\times$8$\times$4$\times$256  \\
    \hline
    Conv-BN-ReLU    & 1$\times$1$\times$1@128  & 8$\times$8$\times$4$\times$128 \\
    \hline
    Conv-BN-ReLU   & 3$\times$3$\times$3@512  & 8$\times$8$\times$4$\times$512  \\
    \hline
    MaxPool    & 2$\times$2$\times$2  & 4$\times$4$\times$2$\times$512  \\
    \hline
    \multicolumn{3}{c}{4$\times$4$\times$2 Global Average Pooling}\\
    \hline
    \multicolumn{3}{c}{3$\times$1 Fully-Connected, Multi-sigmoid } \\
    \multicolumn{3}{c}{3$\times$1 Fully-Connected, Linear   }\\
    \hline \hline
    \end{tabular}
\end{table}
\begin{figure*}
\centering
\includegraphics[trim={250 205 150 125},clip,width=.97\textwidth]{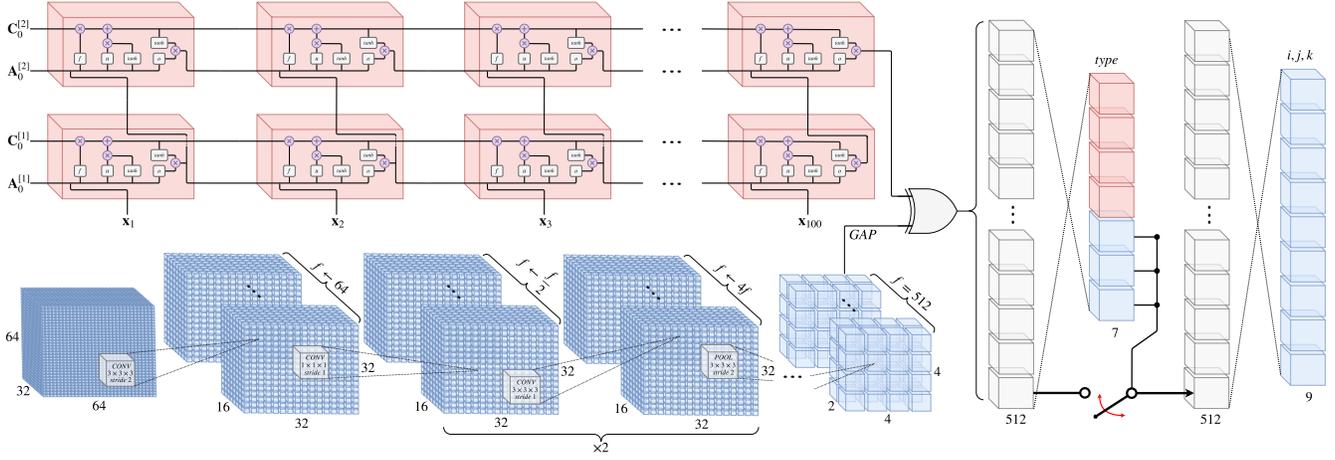}
\caption{Unified framework for time and frequency domain perturbation type classification and coordinate regression. An LSTM network at the top for time domain signals, and a 3D CNN below for frequency domain signals. Both networks output $512$ dimensional latent variable representations of their respective inputs, and their \textit{flow} is controlled by XOR gate logic and a \textit{switch} is activated for perturbation coordinate regression in the frequency domain.}
\label{fig: framework}
\end{figure*}

Table~\ref{table: 3D-CNN_arch} depicts the 3D CNN architecture, devised through experimentation, for the classification of perturbation types in the frequency domain and their respective coordinate locations in 3D space. Convolutional layers use $3\times3\times3$ kernels with stride $1$ and are followed by Batch Normalization (BN)~\cite{ioffe2015batch} and ReLU activations. In order to reduce the number of parameters incurred by 3D convolutions and increase the complexity of the network with more ReLU non-linearities, Bottleneck Layers ($1\times1\times1$ convolution) are introduced between $3\times3\times3$ convolutions. Max Pooling with $2\times2\times2$ kernels down sample inputs and a final Global Average Pooling (GAP)~\cite{lin2013network} layer produces $512$ dimensional vector representations. The representations are then fed to $2$ separate FC Layers, one for multi-label classification with $3$ sigmoid non-linear units and the other for perturbation coordinate regression ($i,j,k$) with $3$ linear units. In the combined perturbation case, the $3$ sigmoid units represent $7$ different classes denoted as
\begin{equation}
\mathbf{C}=\{001,~010,~100,~101,~011,~110,~111\},
\end{equation} 
where $\mathbf{C}$ contains all combinations of Localised (ID $100$), Travelling type~$1$ (ID~$010$) and type~$2$ (ID~$001$) perturbations as described in Section ~\ref{sec: scenarios}. In practice, the $3$ linear units become $9$ units to allow for regression of more than one perturbation location at a time. 

When training a CNN on multiple objectives, it is common practice to compute a linear weighted sum of losses per task $i$ of $T$ tasks, where weight coefficients $\lambda_i$ control the dominance of each loss over the gradient.
Formally, the multi-task optimisation objective is minimised with respect to $\mathbf{W}$ parameters given $\mathcal{D}$ input data as
\begin{equation} \label{eq: multi-objective}
\mathscr{L} = \sum_{i}^{T}{\lambda _i\ell_i(\mathcal{D};\mathbf{W})},
\end{equation}
where $\ell_i$ represents either the negative log-likelihood loss for perturbation type classification: $\ell_{1}(y_1, \widehat{y}_1)$, or the $L2$ loss for perturbation coordinate regression: $\ell_2(y_2, \widehat{y}_2)$. Concretely, the 3D CNN is trained by minimising the following criterion $\mathscr{L}(\mathcal{D}; \mathbf{W}, \lambda_1, \lambda_2)=$
\begin{equation} \label{eq: multi-loss}
\begin{split}
-\frac{1}{N}\sum_{i=1}^{N}\Bigg[&\frac{\lambda_1}{P}\sum_{j=1}^{P}\big[{y^{j}_1\log(\widehat{y}^{j}_{1}) + (1 - y^{j}_1)\log(1 - \widehat{y}^{j}_1)\big]} +\\
 - & \frac{\lambda_2}{C} \sum_{c=1}^{C}{\norm{y^{c}_{2} - \widehat{y}^{c}_{2}}^2} \Bigg]_{i}
\end{split}
\end{equation}
where $P$ and $C$ denote the number of perturbation types and location coordinates respectively, with $\lambda_1$, $\lambda_2$ as tuned weight coefficients for each loss. The resulting network model $\mathcal{F}(\mathcal{D};\mathbf{W})$ predicts a continuous vector of outputs ($i,j,k$~coordinates) and discrete outputs for perturbation type classes. Lastly, parameters $\mathbf{W}$ were initialised as proposed in~\cite{he2015delving}.
\subsection{Time Domain}
Given the sequential nature of the signals in the perturbation induced in the time domain, it was intuitive to utilise Recurrent Neural Networks (RNN). RNNs are particularly suitable for this type of data as their cells can formulate a non linear output $\mathbf{A}^{[t]}$ based on both the input data $\mathbf{x}^{[t]}$ at the current time step $t$, and the previous time-step activation $\mathbf{A}^{[t-1]}$. This is described in~\eqref{eq:RNN}, where $\phi(\cdot)$ is a non-linear activation function of choice such as the hyperbolic tangent.
\begin{equation}
\label{eq:RNN}
\mathbf{A}^{[t]} = \phi(\mathbf{x}^{[t]},\mathbf{A}^{[t-1]})
\end{equation}
In particular, Long Short-Term Memory (LSTM) was adopted because of its capability of learning long term dependencies on data. This is attained by formulating memory cells. The equations relative to LSTM follow, and the reader is invited to refer to the original paper~\cite{hochreiter1997long} for further details.
\begin{equation}
\label{eq:LSTM} 
\begin{split}
& \mathbf{\widetilde{C}}^{[t]}= \tanh(\mathbf{W}_{\tilde{c}} \cdot [\mathbf{A}^{[t-1]},\mathbf{x}^{[t]}]+\mathbf{b}_{\tilde{c}})\\
& \mathbf{\Gamma}_{u}=\sigma(\mathbf{W}_{u}\cdot[\mathbf{A}^{[t-1]},\mathbf{x}^{[t]}]+\mathbf{b}_{u})\\
& \mathbf{\Gamma}_{f}=\sigma(\mathbf{W}_{f}\cdot[\mathbf{A}^{[t-1]},\mathbf{x}^{[t]}]+\mathbf{b}_{f})\\
& \mathbf{\Gamma}_{o}=\sigma(\mathbf{W}_{o}\cdot[\mathbf{A}^{[t-1]},\mathbf{x}^{[t]}]+\mathbf{b}_{o})\\
& \mathbf{C}^{[t]}= \mathbf{\Gamma}_{u} \odot \mathbf{\widetilde{C}}^{[t]} + \mathbf{\Gamma}_{f} \odot \mathbf{C}^{[t-1]} \\
& \mathbf{A}^{[t]} = \mathbf{\Gamma}_{o} \odot \tanh(\mathbf{C}^{[t]})
\end{split}
\end{equation}
In~\eqref{eq:LSTM}, $\mathbf{C}$ is the memory cell, $\mathbf{\Gamma}_{u}$, $\mathbf{\Gamma}_{f}$ and $\mathbf{\Gamma}_{o}$ are the update, forget and output gates respectively; $\mathbf{W}$ denotes the model's weights, and $\mathbf{b}$ are the bias vectors. These parameters are all jointly learnt through backpropagation. Essentially, at each time-step, a candidate update of the memory cell $\mathbf{\widetilde{C}}^{[t]}$ is proposed, and according to the learnt gates, $\mathbf{\widetilde{C}}^{[t]}$ can be utilised to update the memory cell $\mathbf{C}^{[t]}$, and subsequently provide a non linear activation of the LSTM cell $\mathbf{A}^{[t]}$. In order to improve the representational capacity of our network and therefore learn a meaningful representation of the signal, two LSTM layers were stacked, with each LSTM cell containing $512$ neurons. 

The problem of recognising which scenario a signal is representative of was tackled as a multi-label classification task. Since four individual perturbation (and their combinations) were identified (see Table~\ref{table: time_scenarios1-4},~\ref{table: time_scenarios5-6} and~\ref{table: time_scenarios7-15}), in order to classify which of these perturbation was present, $512$ dimensional LSTM representations were fully connected to four neurons with sigmoid activation functions. During training the following negative log-likelihood criterion was minimised
\begin{equation} \label{eq: cross-entropy-loss}
\begin{split}
\mathscr{L}(y, \widehat{y})=-\frac{1}{PN}\sum_{j=1}^{P}\sum_{i=1}^{N}\Big[y_{j} & \log (\widehat{y}_{j}) + \\ 
& (1 - y_{j})\log(1 - \widehat{y}_{j})\Big]_{i}
\end{split}
\end{equation}
where $P$ is the number of sigmoid units used for the multi-label classification task, and $N$ is the number of samples in a batch. The parameters of the resulting model were initiliased as per the scheme proposed in~\cite{glorot2010understanding}.
\begin{figure}
\centering
\includegraphics[trim={120 40 120 10},clip,width=0.95\columnwidth]{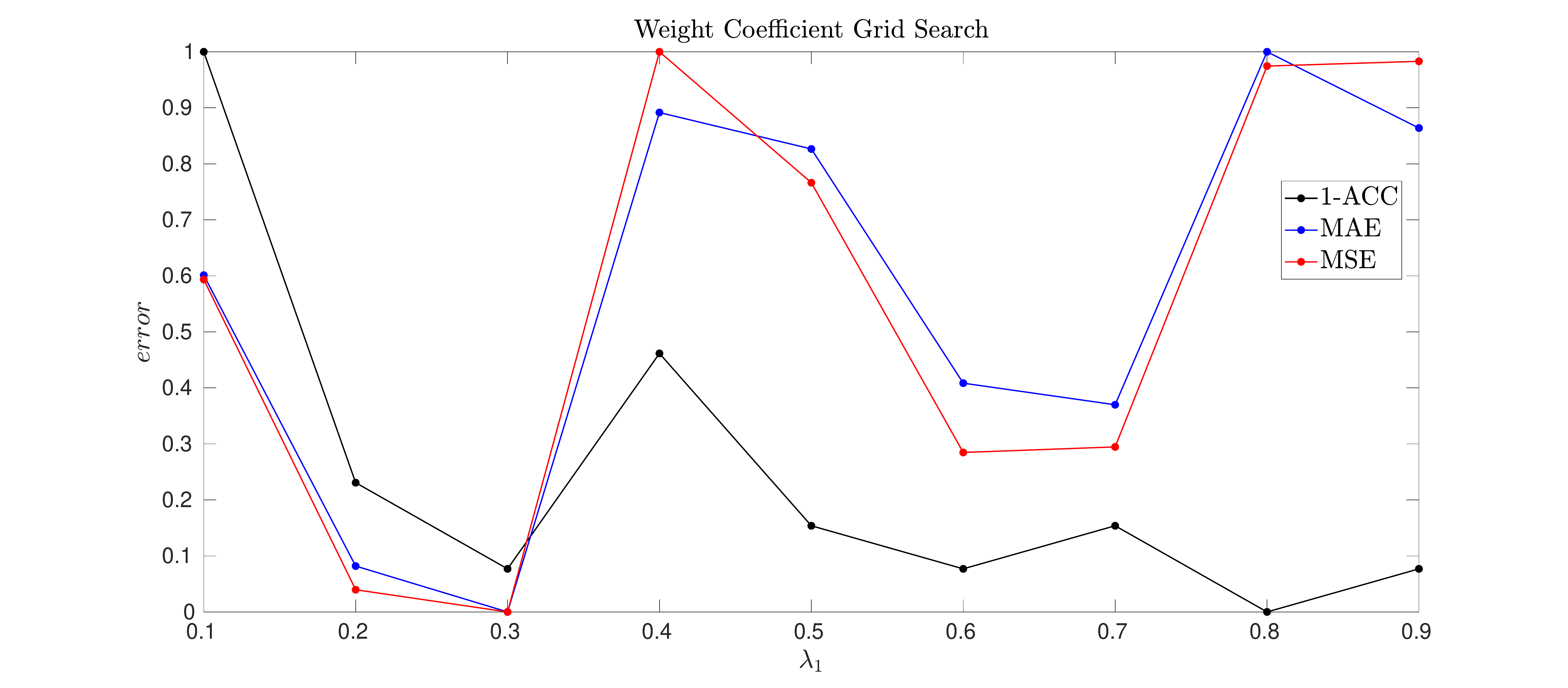}
\caption{Weight coefficient grid search for the 3D-CNN classification and regression losses. Coefficient $0.3$ for classification and $0.7$ for regression yielded the best performance.}
 \label{fig:grid search}
\end{figure}
\subsection{Deep Time-Frequency Framework}
As illustrated in Fig.~\ref{fig: framework}, a Deep Neural Network (DNN) framework was formulated for processing both Time and Frequency Domains signals coming from nuclear reactor sensor measurements. It is important to clarify that the simulations were performed using different rector cores in each domain, as per CORE-SIM and S3K specifications. Both the 3D CNN and LSTM network produced $512$ dimensional vector representations of their respective inputs. The representations were then fed to a fused classification layer comprised of $7$ sigmoid units ($3$ for Frequency \& $4$ for Time) to accommodate all scenario combinations as a multi-label classification task. Lastly, whenever a frequency domain perturbation is detected, the red switch in Fig.~\ref{fig: framework} is triggered and the current $512$ dimensional representation is fed to a separate FC layer to regress perturbation coordinates ($i,j,k$) in 3D space.
\begin{table}
\caption{Results of the frequency domain 3D CNN experiments for perturbation type classification and localisation regression. (*) marks combined perturbations scenarios.}
    \label{table: 3D-CNN}
    \centering
    \setlength{\extrarowheight}{2pt}
    \begin{tabular}{c|c|c|c|c}
    \hline  \hline
    \multicolumn{5}{c}{3D CNN Perturbation Classification \& Localisation} \\[.05cm]
    \hline
    Sensors & Train/Valid/ & Classification & \multicolumn{2}{c}{($i,j,k$) Regression} \\[.01cm]\cline{4-5}
    (\%)& Test (\%) & Accuracy (\%) & MAE & MSE \\[.01cm]
    \hline
	20 & 60/15/25 & \textbf{99.75$\pm$0.09} & \textbf{0.2528$\pm$0.03} & \textbf{0.1347$\pm$0.02} \\[.01cm]
	20 & 25/15/60 & 99.12$\pm$0.17 & 0.4221$\pm$0.05 & 0.4152$\pm$0.07 \\[.01cm]
 	20 & 15/25/60 &	98.62$\pm$0.22 & 0.5886$\pm$0.05 & 0.8174$\pm$0.12 \\[.01cm]
 	\hline
	5 & 60/15/25 &	99.32$\pm$0.18 &	0.326$\pm$0.05 &	0.2086$\pm$0.04	\\[.01cm]
    5 &	25/15/60 &	98.34$\pm$0.22	& 0.4818$\pm$0.05 &	0.6044$\pm$0.08 \\[.01cm]
    5 &	15/25/60 &	97.27$\pm$0.54 &	0.689$\pm$0.1 &	1.0749$\pm$0.25 \\[.01cm]
    \hline
    20* & 60/15/25 &	99.82$\pm$0.05	& 0.5602$\pm$0.04 &	1.6036$\pm$0.15 \\[.01cm]
    20* & 25/15/60 &	99.56$\pm$0.07 &	0.8942$\pm$0.04 &	3.5739$\pm$0.16 \\[.01cm]
    20* & 15/25/60 &	99.44$\pm$0.08 &	0.9635$\pm$0.06 &	4.2814$\pm$0.19 \\[.01cm]
    \hline
 	5* & 60/15/25 & 99.47$\pm$0.03 & 0.8809$\pm$0.04& 3.4424$\pm$0.16\\[.01cm]
    5* &	25/15/60 & 98.33$\pm$0.24 & 0.5001$\pm$0.04 & 0.6381$\pm$0.08 \\[.01cm]
    5* & 15/25/60 & \textbf{97.15$\pm$0.15} & \textbf{1.9528$\pm$0.11} & \textbf{11.902$\pm$0.66} \\[.01cm]
    \hline \hline
    \end{tabular}
\end{table}
\section{Experimental Study}
\subsection{Frequency Domain}
For completeness and more detailed analysis of the results, the performance of the proposed framework in the Time and Frequency domains are kept separate. The implementation was based on MATLAB~\cite{guide1998mathworks}, Keras deep learning framework~\cite{chollet2015keras} and Tensorflow numerical computation library~\cite{abadi2016tensorflow}. The experiments were conducted using a server with an Intel Xeon(R) E5-2620 v4 CPU, eight GPUs and 96GB of RAM. The results of the experiments conducted on the volumetric signal data are reported in Table~\ref{table: 3D-CNN}. As explained in greater detail in Subsection~\ref{subsec: data-preprocessing}, the volumetric signals were corrupted by obscuring parts at random, in order to emulate fewer available sensor measurements and thus increase the complexity of the problem. As shown in Table~\ref{table: 3D-CNN}, in the first experiment a dataset with $20$\% of the sensor measurements was generated. Similarly, in the second experiment a different dataset was generated in which only $5$\% of the sensor measurements were kept. Both of these experiments were conducted to study the effect of sensor measurement sparsity on the performance of our algorithm. Furthermore, different training, validation and test splits were also utilised to study the effect of learning from a smaller pool of possible perturbations in the training set. 

In the case of the Combined Perturbation experiments (marked with (*) in Table~\ref{table: 3D-CNN}), a similar approach was undertaken with regards to the percentage of sensors kept and the dataset splits. Two datasets were generated for training ($20$\% and $5$\%) in which multiple perturbations are classified and their respective source coordinates regressed simultaneously. Moreover, a hyper-parameter grid search was performed over the loss weight coefficients for each task, and the best results were achieved with $\lambda_1 = 0.3$ and $\lambda_2 = 0.7$ (see Fig.~\ref{fig:grid search}). For all experiments, the 3D CNN was trained to minimise the criterion in Eq.~\eqref{eq: multi-loss} using backpropagation, and the Adaptive Moment Estimation (Adam) optimiser~\cite{kingma2014adam} with the default parameters and a batch size of size $32$. Each model was trained $10$ times and the mean performance was taken as the final result, along with the standard deviation. 

As observable in Table~\ref{table: 3D-CNN}, high classification performance was achieved in all experiments with $99.75$\%$\pm0.09$ and $97.15$\%$\pm0.15$ accuracy in the best case and worst case respectively. The mean squared and absolute errors (MSE, MAE) were used as evaluation metrics for the perturbation coordinate regression results, with best case of $0.2528\pm0.03$ (MAE), $0.1347\pm0.02$ (MSE) and worst case of $1.95\pm0.11$ (MAE), $11.90\pm0.66$ (MSE). 
\begin{figure}[t]
\centering
\includegraphics[trim={0 0 178cm 0},clip,width=.97\columnwidth]{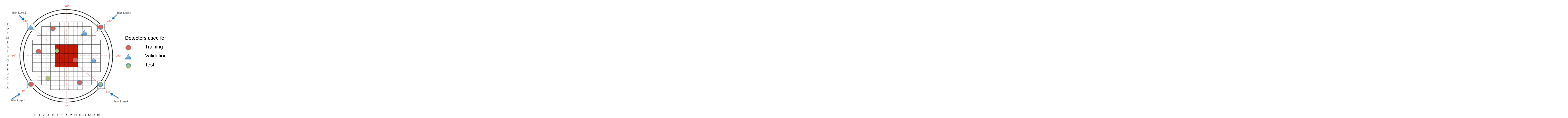}
\caption{Description of detector locations for the signals utilised in training, validation and testing of the deep LSTM network model, in the classification of different types and combinations of Time domain perturbations.}
\label{fig:reactor_core_sensors_train_valid_test}
\end{figure}

Overall, the results show that the classification task achieves better performance across all datasets compared to the regression task. The regression performance deteriorates with the introduction of combined perturbations and limited sensor measurement/training set size, whereas the classification of perturbations types is more resilient to fluctuations in the number of sensors used in the training phase.
\subsection{Time Domain}
\begin{figure}
\centering
\includegraphics[trim={20 20 10 2},clip,width=.99\columnwidth]{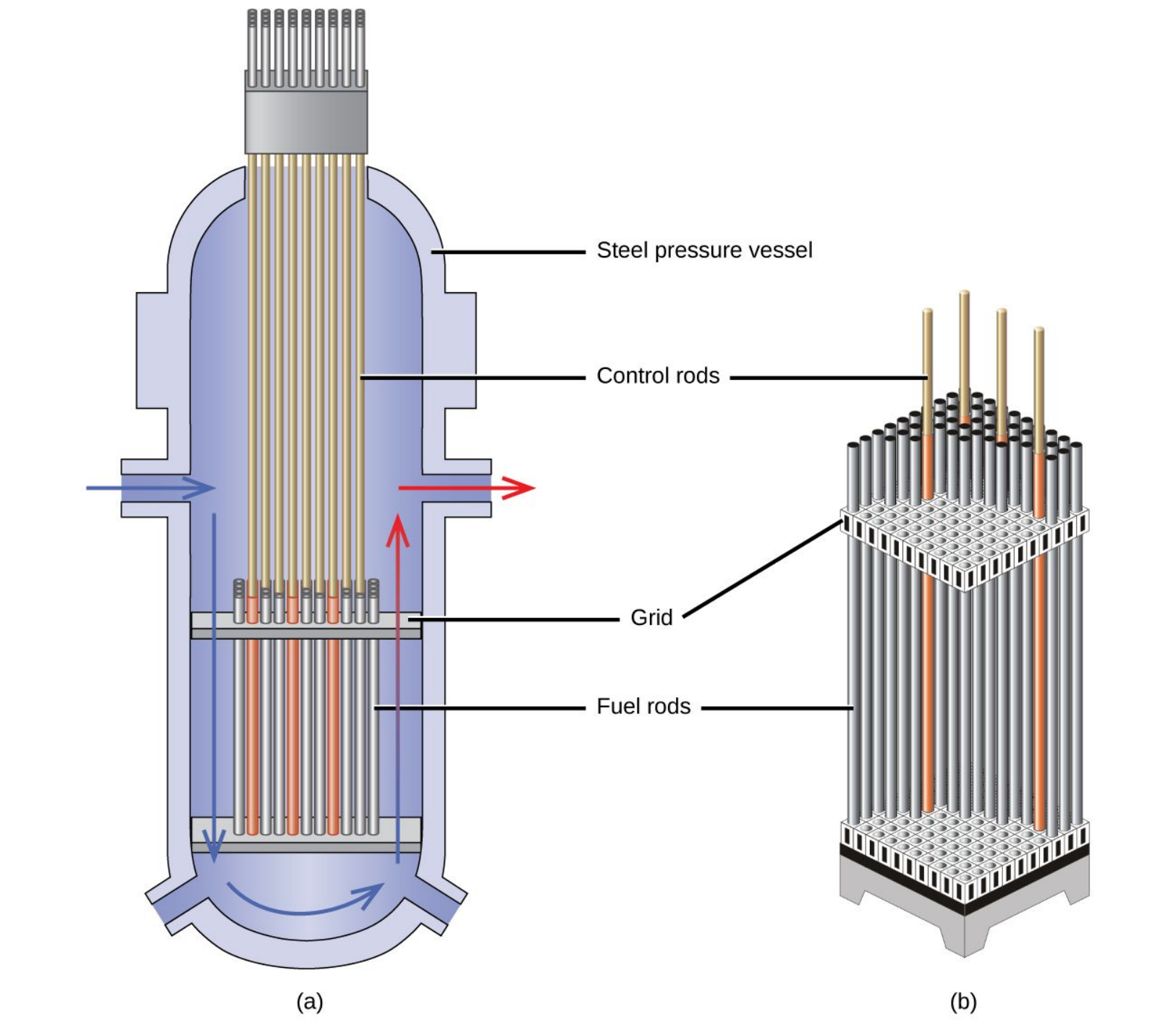}
\caption{Illustration of a nuclear reactor core, highlighting its internal components. The core shown on the left hand side contains the fuel and control rod assembly shown on the right hand side. Photo credit to~\cite{reactor2018,reactorcomponents2018}.}
\label{fig: fuel_assembly}
\end{figure}

In this experiment, individual sensor measurements were utilised to detect each of the thirteen scenarios (Table~\ref{table: time_scenarios1-4},~\ref{table: time_scenarios5-6} and~\ref{table: time_scenarios7-15}). Starting with the data from the thirteen scenarios provided by S3K, each comprised of $56$ one-dimensional signals of length $10001$ (one signal per detector), after re-sampling, $17164$ samples of size $56\times 100$ were obtained. Subsequently, each $56\times 100$ sample was subdivided into $56$ one-dimensional signals of size $100\times 1$. During training, each $100\times 1$ signal from a single sensor was utilised to detect the presence of a scenario. In other words, any given scenario (perturbation) must be detected within one second of monitoring. Fig.~\ref{fig:reactor_core_sensors_train_valid_test} shows which sensors were utilised to train, validate and test the LSTM network, at each radial level of the reactor. For better intuition, Fig.~\ref{fig: fuel_assembly} provides a depiction of the main components of a core, including fuel assemblies in a typical nuclear reactor. Overall, $480592$ (from $28$ sensors), $240296$ (from $14$ sensors) and $240296$ (from $14$ sensors) signals were used for training, validating and testing.

Hyper-parameters were experimentally tuned, and those utilised provided the best performance. The negative log-likelihood criterion in~\eqref{eq: cross-entropy-loss} was minimised with mini-batch ($32$) stochastic gradient descent (SGD). The Adam optimisation algorithm was used, to include adaptive learning rate, momentum, RMSprop and bias correction in weight updates, offering faster convergence rate than normal SGD with momentum~\cite{kingma2014adam}. The classification accuracy (\%) achieved by the LSTM network was $97\%$ on the clean signals, $81\%$ with added noise at SNR$=10$ and $77\%$ with added noise at SNR$=5$. 
\section{Conclusion \& Future Work}
In this paper, the first step towards a unified deep framework was proposed for the classification and regression of perturbations in nuclear reactors. Both Time and Frequency domain data were obtained through inducing perturbations such as an \textit{Absorber of Variable Strength} and \textit{Propagating Perturbation} in the Frequency domain; vibration of fuel assemblies and fluctuations of thermal-hydraulic parameters at the inlet coolant between the 4-loops of a Westinghouse PWR reactor in the Time domain. 

The proposed framework is comprised of a 3D CNN and an LSTM architecture that each output $512$ dimensional representations of their respective input signals, and combinations of nuclear reactor perturbations are classified with a fused multi-sigmoid layer. A \textit{switch} was introduced to control the \textit{flow} of the frequency domain $512$ dimensional representation, which is fed to a regression layer whenever a perturbation is detected in the 3D complex signal volume. Furthermore, the effects of sensor measurement sparsity and noisy signals were evaluated in a series of experimental studies, demonstrating the capability of our framework to achieve good results in both unfolding and perturbation type classification.   

In future work, we plan to extend our studies to other types of data, simulated in the Time and Frequency domains utilising the same/multiple reactor cores, to test the sensitivity of our framework to different reactor characteristics. Furthermore we intend to investigate real data coming from nuclear power plants, in pursuit of a framework suitable for simultaneously handling Time and Frequency domain signals for the localisation and classification of nuclear reactor anomalies. 

\section*{Acknowledgment}
The research conducted was made possible through funding from the Euratom research and training programme 2014-2018 under grant agreement No~754316 for the `CORe Monitoring Techniques And EXperimental Validation And Demonstration (CORTEX)' Horizon 2020 project, 2017-2021. We would like to thank the reviewers for their helpful comments.

\bibliographystyle{unsrt}
\bibliography{main}

\end{document}